\documentclass[external]{esp-paper} 
\usepackage{multirow}
\usepackage{bm}
\usepackage{hyperref}
\usepackage{cleveref}
\usepackage{multirow}
\usepackage{enumitem}

\title{Beyond the Baseband: \\ Adaptive Multi-Band Encoding \\ for Full-Spectrum Bioacoustics Classification}

\author{Eklavya Sarkar$^{\star}$}
\author{Marius Miron}
\author{David Robinson}
\author{Gagan Narula}
\author{Milad Alizadeh}
\author{Ellen Gilsenan-McMahon}
\author{Felix Effenberger}
\author{Emmanuel Chemla}
\author{Olivier Pietquin}
\author{Matthieu Geist}
\affil{Earth Species Project}

\begin{document}
\thispagestyle{firstpage}
\maketitle

\begingroup
\def\thefootnote{$\star$}\footnotetext{Corresponding author: \url{eklavya@earthspecies.org}}

\def\thefootnote{\arabic{footnote}}
\endgroup

\begin{abstract} 
Animals hear and vocalize across frequency ranges that differ substantially from humans, often extending into the ultrasonic domain. Yet most computational bioacoustics systems rely on audio models pre-trained at 16 kHz, restricting their usable bandwidth to the 0-8 kHz baseband and discarding higher-frequency information present in many bioacoustic recordings. We investigate a multi-band encoding framework that decomposes the full spectrum of animal calls into band features and fuses them into a unified representation. Similarity analyses on models show that certain encoders produce decorrelated band embeddings that improve class separation after fusion. Classification experiments on three bioacoustic datasets using eight pre-trained models and five fusion strategies show that fused representations consistently outperform the baseband and time-expansion baselines on two datasets, showing the potential of multi-band methods for full-spectrum encoding of animal calls.
\end{abstract}

\section{Introduction}
Bioacoustics, the study of animal sounds, has emerged as a prominent domain within machine learning, as a means to study the origins and evolution of language and vocal learning \citep{hurford2012language, fitch2018biology}, and deepen our understanding of communication in the natural world. Computational bioacoustics aims to `decode' animal vocalizations to gain insights into their communication by deriving information from their calls \citep{bioacoustics_roadmap}. In recent years, neural representations learnt on human speech have shown remarkable transferability to the domain of bioacoustics for decoding non-human vocal communication \citep{Sarkar_Thesis_2025, cauzinille25_phd, miron2026what}. Indeed, modern state-of-the-art foundation models pre-trained on human speech and/or general audio have achieved strong performance benchmark across a range of species for tasks such as call detection \citep{BEANS, aves, semenzin2025dolphvec}, call-type classification \citep{kloots24_vihar, shi24_vihar, mahoud24_vihar, abzaliev24, Sarkar_ICASSP_2025}, or caller identification \citep{sarkar23_interspeech, cauzinille24_interspeech, Knight2024}, thus providing a reliable framework for bioacoustics tasks, and significantly advancing the field.

\begin{figure}[htb]
  \centering
  \includegraphics[width=\linewidth]{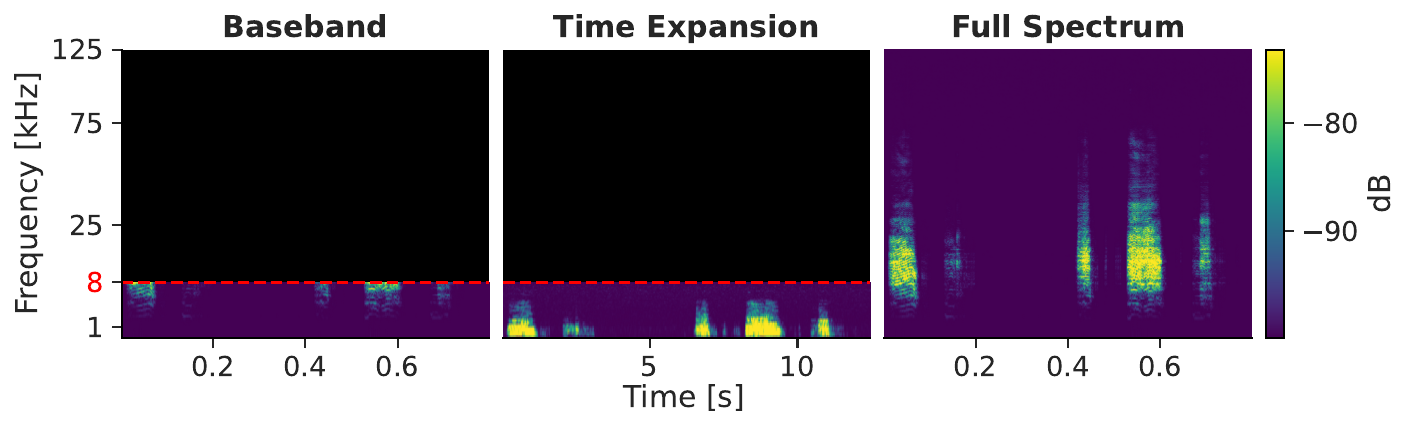}
  \caption{Spectrograms of an ultrasonic bat vocalization.}
  \label{fig:cutoff}
\end{figure}

However, a critical limitation in leveraging these models is that they are typically pre-trained at a 16 kHz sampling rate (SR), corresponding to the \textit{human} audible range, and resample any input to the constrained 0--8 kHz baseband, including bioacoustic recordings. As depicted in \Cref{fig:cutoff}, this bandwidth (BW) is highly insufficient to cover the higher frequency content contained in the full-spectrum of many animals and their vocalizations, which extend well beyond this threshold, and results in a significant loss of useful information. For example, bat echolocation calls can reach up to 200 kHz \citep{bats_altringham}, insect signals to 100 kHz \citep{drosopoulos2005insect}, and marine mammal calls to 150 kHz \citep{berta2005marine}. However, training entire foundation models at higher SRs is very computationally expensive, and only a few such models exist, such as BirdNET \citep{kahl2021birdnet} and Perch \citep{Ghani2023, perch2}, pre-trained at 48 and 32 kHz, respectively, on bioacoustics. Thus, the fixed pre-training bandwidth of existing audio models is a major limitation in all current computational bioacoustics works.

A common signal-processing approach to address this problem is time-expansion, i.e. slowing down the audio recordings by a fixed factor to shift the high-frequency components down to the baseband of the pre-trained model \citep{preatoni2005identifying, kershenbaum2025automatic, bats2000}, and using the resulting signal as input. However, this approach considerably reduces the spectral resolution and proportionally stretches the recording, as illustrated in \Cref{fig:cutoff} (middle), thus increasing the model's computational cost of processing the signal.

Inspired by classical sub-band processing approaches in automatic speech recognition (ASR), where the signal is decomposed into frequency bands that are processed independently before combining their evidence to improve robustness \citep{bourlard1996mew, bourlard1996multi, tibrewala1997multi, hermansky1998traps}, we investigate an adaptive multi-band (MB) encoding strategy for bioacoustics. Unlike in ASR, where such methods operated on sub-bands within the 0--8 kHz baseband, we instead decompose the full spectrum of animal vocalizations into multiple basebands, compatible with modern pre-trained audio models, and explore combining them into a unified representation through various fusion strategies. Based on this proposed approach, this paper investigates the following two central questions:
\begin{enumerate}[leftmargin=*]
    \item Can MB representations effectively exploit the unused high-frequency information in bioacoustic calls, and improve over the conventional baseband and time-expansion approaches? 
    \item How does this approach compare to simply using the baseband of a model trained at a higher SR, such as BirdNET, at 48 kHz? Moreover, when applied to such a model, does it yield additional gains over its application to standard 16 kHz models?
\end{enumerate}

To answer these questions, we conduct a systematic evaluation across multiple bioacoustics datasets and pre-trained audio models. It is to be noted that we do not provide a foundation model pre-trained for higher SRs in this paper, but instead propose an approach that can work with any existing model. To provide practical use to the bioacoustics community, we also release the framework as an open-source toolkit.\footnote{Source code: \url{https://github.com/earthspecies/multiband-audio}.}

The rest of this paper is organized as follows. \Cref{sec:method} details the MB encoding approach, and \cref{sec:setup} gives our experimental setup. \Cref{sec:rep_analysis}--\ref{sec:results} respectively present a representation and classification analysis. \Cref{sec:conclusion} concludes the paper.

\section{Adaptive Multi-Band Encoding}
\label{sec:method}
\begin{figure*}[!htb]
  \centering
  \includegraphics[width=\linewidth]{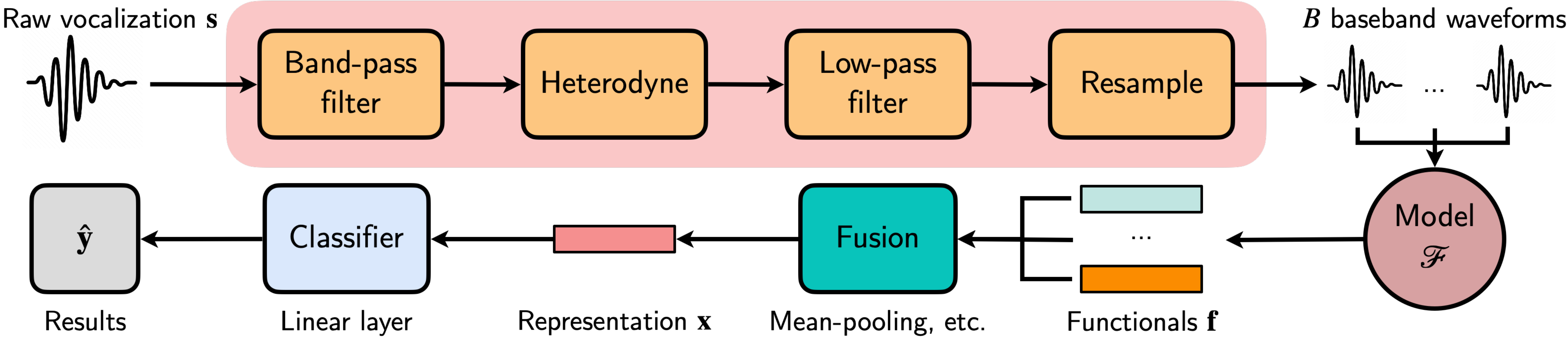}
  \caption{Complete pipeline of the heterodyning-based multi-band encoding.}
  \label{fig:pipeline}
\end{figure*}

This section details the heterodyning-based multi-band processing approach, illustrated in \Cref{fig:pipeline}. It consists of three main stages: (i) spectral band decomposition and baseband mapping, (ii) per-band representation extraction using a frozen pre-trained encoder, and (iii) fusion of the resulting band-level representations into a unified embedding for downstream classification. 

Given a raw audio recording $\mathbf{s}$ sampled at its native rate $f_s$, a pre-trained model operating at $f_m$ can only access frequencies up to its Nyquist $f_{m}/2$. To leverage the higher spectral information $\mathbf{s}$, we split its full spectrum into $B = \lceil f_s / f_m \rceil$ non-overlapping bands of width $f_{m}/2$. The first band ($0$--$f_{m}/2$ Hz) corresponds to the standard baseband, and is not further processed. For each subsequent band $b$, we first extract the corresponding spectral slice by applying a band-pass filter to the raw signal $\mathbf{s}$, yielding $\mathbf{s}_b$ centered at frequency $f_b$. We then heterodyne it down to the baseband by multiplying it by a cosine at the band's center frequency $f_b$, followed by low-pass filtering $\mathcal{H}_{\mathrm{LP}}$. The resulting baseband signal $\bar{\mathbf{s}}_b$ is given by:
\begin{equation*}
\bar{\mathbf{s}}_b
= \mathcal{H}_{\mathrm{LP}}\bigl[\mathbf{s}_b\cdot\cos(2\pi f_b t)\bigr]
\end{equation*}

Applying this to each band produces $B$ baseband waveforms $\bar{\mathbf{S}}=\{\bar{\mathbf{s}}_1, \ldots, \bar{\mathbf{s}}_B\}$, each representing a distinct portion of the original spectrum. We resample them to $f_m$, matching the SR expected by the pre-trained model, and then pass them individually through the frozen encoder $\mathcal{F}$, yielding $B$ variable-length multi-dimensional embeddings. These are then transformed into $B$ fixed-length vocalization-level functionals $\mathbf{f}_b\in\mathbb{R}^{D}$ by computing the first-order statistics across the temporal axis, where $D$ denotes the embedding dimension of the model. Finally, a learned fusion module combines these functionals into a final representation $\mathbf{x} \in \mathbb{R}^D$, on which a linear classifier is trained for the downstream task. In this paper, we investigate five distinct fusion strategies, detailed below:

\begin{itemize}[itemsep=0em, leftmargin=*]
    \item \textbf{Mean-Pool} (MP): $\mathbf{f}$ are averaged element-wise across the band dimension, assigning equal weight coefficients to all frequency bands. No new learnable parameters are introduced.
    \item \textbf{Gated-Pool} (GP): A linear projection maps each $\mathbf{f}_b$ to a scalar score, which is normalized via softmax to produce band-level weights coefficients $w_b$. The fused representation is the weighted sum of the band functionals $\mathbf{x} = \sum_b w_b \mathbf{f}_b$, allowing the model to learn which frequency bands are most informative.
    \item \textbf{Mixture-of-Experts} (MoE): Each $\mathbf{f}_b$ is passed through an independent linear classifier head to obtain band-level logits $z_b$. A separate 2-layer MLP computes band-level weights coefficients from $\mathbf{f}_b$. The final prediction is the weighted sum of band logits $\mathbf{\hat{y}} = \sum_b w_b z_b$. Unlike GP, which fuses $\mathbf{f}$ before classification, MoE classifies them first and fuses the resulting logits.
    \item \textbf{Hybrid} (HYB): Similar to GP, but the gating network additionally incorporates handcrafted features, namely spectral entropy and flux, alongside each $\mathbf{f}_b$ to compute band weight coefficients. The concatenated representation is input to a 3-layer MLP, enabling the gating decision to leverage both the learned functionals and acoustic properties of each band.
    \item \textbf{Self-Attention} (SA): $\mathbf{f}$ are treated as a sequence of tokens, with a learnable [CLS] token and positional embeddings. A single-layer transformer encoder processes the sequence, contextualizing each band's representation with all other bands through a multi-head attention mechanism. The [CLS] output serves as the fused representation $\mathbf{x}$.
\end{itemize}

\section{Experimental Setup}
\label{sec:setup}

\subsection{Datasets, Tasks, and Protocols}
We investigate our method on three distinct bioacoustic datasets (DS) in the BEANS \citep{BEANS} benchmark, summarized in \Cref{table:dataset_stats}.
\begin{table}[!htb]
\centering
\caption{SR is given in kHz. $S$ is the \# of samples, $L$ length [mins.], $n_c$ \# of classes, $\mu$ median length [s], $\sigma$ std, and $B$ the number of 8 kHz bands.}
\begin{tabular}{lrrrrrrr}
\toprule
\textbf{DS} & \textbf{SR} & $\bm{S}$ & $\bm{L}$ & $\bm{n_c}$ & $\bm{\mu}$ & $\bm{\sigma}$ & $\bm{B}$\\
\midrule
Dogs & $44.1$ & $688$ & $126$ & $10$ & $8.05$ & $16.01$ & 3\\
CBI  & $44.1$ & $21$K & $3351$ & $264$ & $10.00$ & $1.70$ & 3\\
Bats & $250$ & $10$K & $313$ & $10$ & $1.45$ & $1.08$ & 16 \\
\bottomrule
\end{tabular}
\label{table:dataset_stats}
\end{table}

The downstream tasks consist of caller identity classification for Dogs and Bats, and species classification for Cornell Birdcall Identification (CBI). We follow the protocols defined in BEANS to split the datasets into \textit{Train}, \textit{Val}, and \textit{Test} sets.

\subsection{Models, Feature Representations, and Baselines}
For our work, we consider different families of pre-trained (PT$_0$) models to obtain distinct features $\mathcal{F}$. They are post-trained (PT$_1$) on various dataset permutations, given in \Cref{table:models}.

\begin{table}[ht]
\centering
\caption{\# Parameters $P$ [M] and feature dimension $D$ of models. IN represents ImageNet, AS AudioSet, XC Xeno-Canto.} 
\begin{tabular}{lllllll}
\toprule
\bm{$\mathcal{F}$} & \textbf{PT$_0$} & $\mathbf{{DS}_0}$ & \textbf{PT$_1$} & $\mathbf{{DS}_1}$ & $\bm{P}$ & $\bm{D}$ \\
\midrule
 EffNet-Bio & SL & IN   & SL & Bio   & 5  & 1280\\
 EffNet-AS  & SL & IN   & SL & AS    & 5  & 1280\\
 EffNet-All & SL & IN   & SL & All   & 5  & 1280\\
\midrule
BEATs-Bio   & SSL & AS & SL & Bio & 91 & 768\\
BEATs-All   & SSL & AS & SL & All & 91 & 768\\
BEATs-NLM & SSL & AS & ALM  & Misc   & 91 & 768\\
\midrule
EATs-All & SSL & AS & SL & Bio & 90 & 768 \\
EATs-Bio & SSL & AS & SL & Bio & 90 & 768\\
\midrule
BirdNET & SL & XC & -- & -- & 14 & 1024 \\
\bottomrule
\end{tabular}
\label{table:models}
\end{table}

\textbf{SL pre-trained on ImageNet}: We select \textit{EfficientNet} for its moderate size and strong performance on BEANS \citep{miron2026what}. It is a 16 kHz CNN model pre-trained on ImageNet. We evaluate three variants\footnote{\url{https://github.com/earthspecies/avex}.} post-trained on bioacoustics data (\textit{Bio}), AudioSet (\textit{AS}), or both (\textit{All}) \citep{miron2026what} to assess the impact of the post-training domain for leveraging higher-frequency content. We also include \textit{BirdNET}, a 48 kHz model based on the same architecture and post-trained primarily on Xeno-Canto (XC), to directly compare performance across BWs and test if our multi-band approach generalizes to models with higher native SRs.

\textbf{SSL pre-trained on speech and general audio}: \textit{BEATs} is a larger model composed of a CNN module operating on mel-spectrogram input patches, followed by 12 transformer layers. We pick two variants post-trained on \textit{Bio} and \textit{All}, and Nature-LM audio's BEATS encoder, extracted from an audio-language model (ALM) trained on bioacoustic audio and text pairs \citep{naturelm}.

\textbf{SSL pre-trained on human speech, general audio, and bioacoustics}:
\textit{EATs} is a SSL pre-trained with teacher distillation and masked spectrogram reconstruction tasks on \textit{Bio}, \textit{AS}, and \textit{All} sets. We select the variants post-trained on \textit{Bio} and \textit{All}.

For all the models, we extract the final encoder layer. We compare our multi-band approach with the following baselines.
\begin{itemize}[itemsep=0em, leftmargin=*]
    \item \textbf{Baseband} (BB): we simply resample $\mathbf{s}$ to the given model's baseband, and lose any spectral content above this threshold. 
    \item \textbf{Time-Expansion} (TE): we slow down $\mathbf{s}$ to shift and compress all the spectral information down to the model's baseband.
\end{itemize}

\section{Representation Analysis}
\label{sec:rep_analysis}

This section presents similarity analysis of the extracted features $\mathbf{x}$. For these studies we only work with the \textit{Train} set.

\subsection{Band Discrimination}
\label{ssec:band_disc}
\begin{figure}[!htb]
  \centering
  \includegraphics[width=\linewidth]{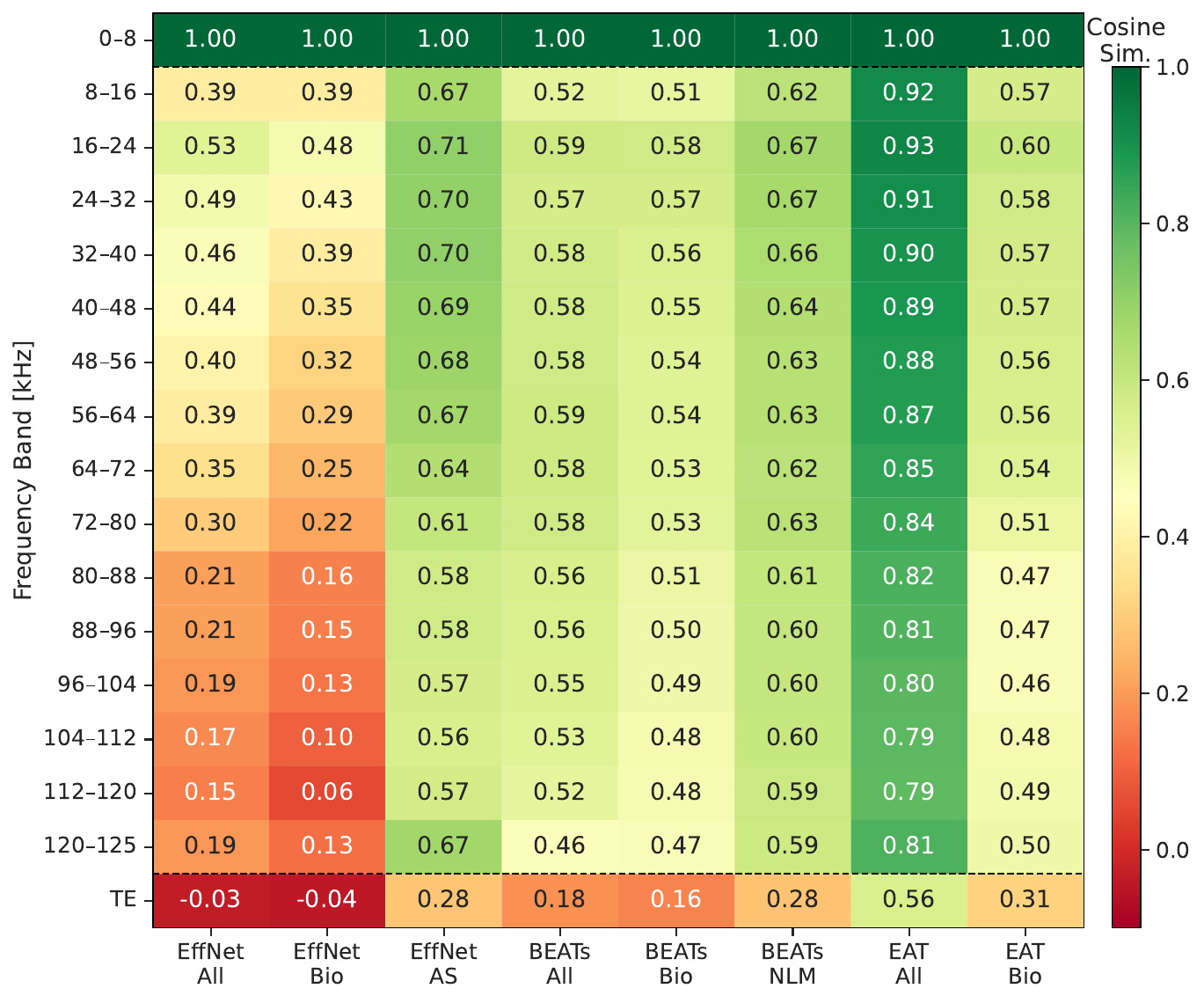}
  \caption{Mean cosine similarity between baseband and other band's embeddings for Bats. Higher (greener) values indicate closer representational similarity with the baseband.}
  \label{fig:bandwise_bats}
\end{figure}

We first explore how the extracted features vary across frequency bands, given that none of the models were exposed to high-frequency content during PT$_0$ or PT$_1$. To that end, \Cref{fig:bandwise_bats} shows the mean cosine similarity between the baseband (0–8 kHz) and the embeddings extracted from all the other frequency bands, post-heterodyning, for the Bats dataset. For EffNet-All and Bio, we can notably observe that the similarity decreases progressively with higher frequency bands, indicating that they produce representations increasingly decorrelated from the baseband, which could benefit the fusion methods by adding complementary information. In contrast, EAT-All maintains consistently high similarity across the frequency spectrum, implying that features from different bands are encoded very similarly to the baseband. The remaining models also only show moderate similarity with the baseband, with some variation across bands. TE yields near-zero or negative scores, showing that compressing the entire 125 kHz spectrum into the 8 kHz baseband via a 15.6x time-expansion produces fundamentally different representations. Dogs and CBI features yield similar overall patterns, suggesting similar behavior for band similarity.

\subsection{Class Separation}
\begin{figure}[!htb]
  \centering
  \includegraphics[width=\linewidth]{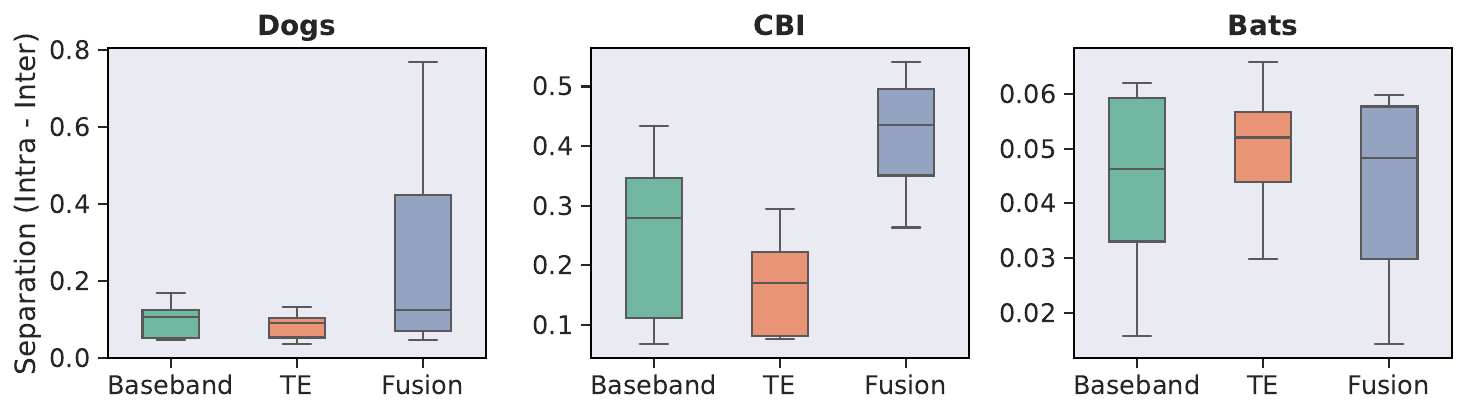}
  \caption{Class separation per method. Higher values indicates more discriminative embeddings.}
  \label{fig:cos_distances}
\end{figure}

We also investigate the class discrimination of the extracted representations. To this end, we compute the pairwise mean intra and inter-class cosine similarities, and measure their separation as the difference between the two. In an ideal scenario, intra-class similarity is high and inter-class similarity is low, yielding a high class separation value. \Cref{fig:cos_distances} visualizes the resulting class-separation scores for each approach, aggregated across all models. The fusion distributions are taken from all strategies.

For Dogs, fusion shows high variance across models. The median separation is comparable to BB and TE, but some models achieve substantially higher scores, indicating that fusion can yield more discriminative representations for certain models. For CBI, fusion shows a clear improvement over both methods. Finally, for Bats, all methods show comparable separation, with TE achieving a marginally higher median. As shown in \Cref{fig:bandwise_bats}, most models encode higher frequency bands similarly to the baseband, meaning fusion combines largely redundant representations. In contrast, TE produces near-orthogonal features that yield slightly better class separation.

\section{Classification Results}
\label{sec:results}

This section analyzes the downstream classification performance of the extracted features, using a linear head trained for 20 epochs. We evaluate the performance with accuracy on \textit{Test}.

\begin{figure}[!htb]
  \centering
  \includegraphics[width=\linewidth]{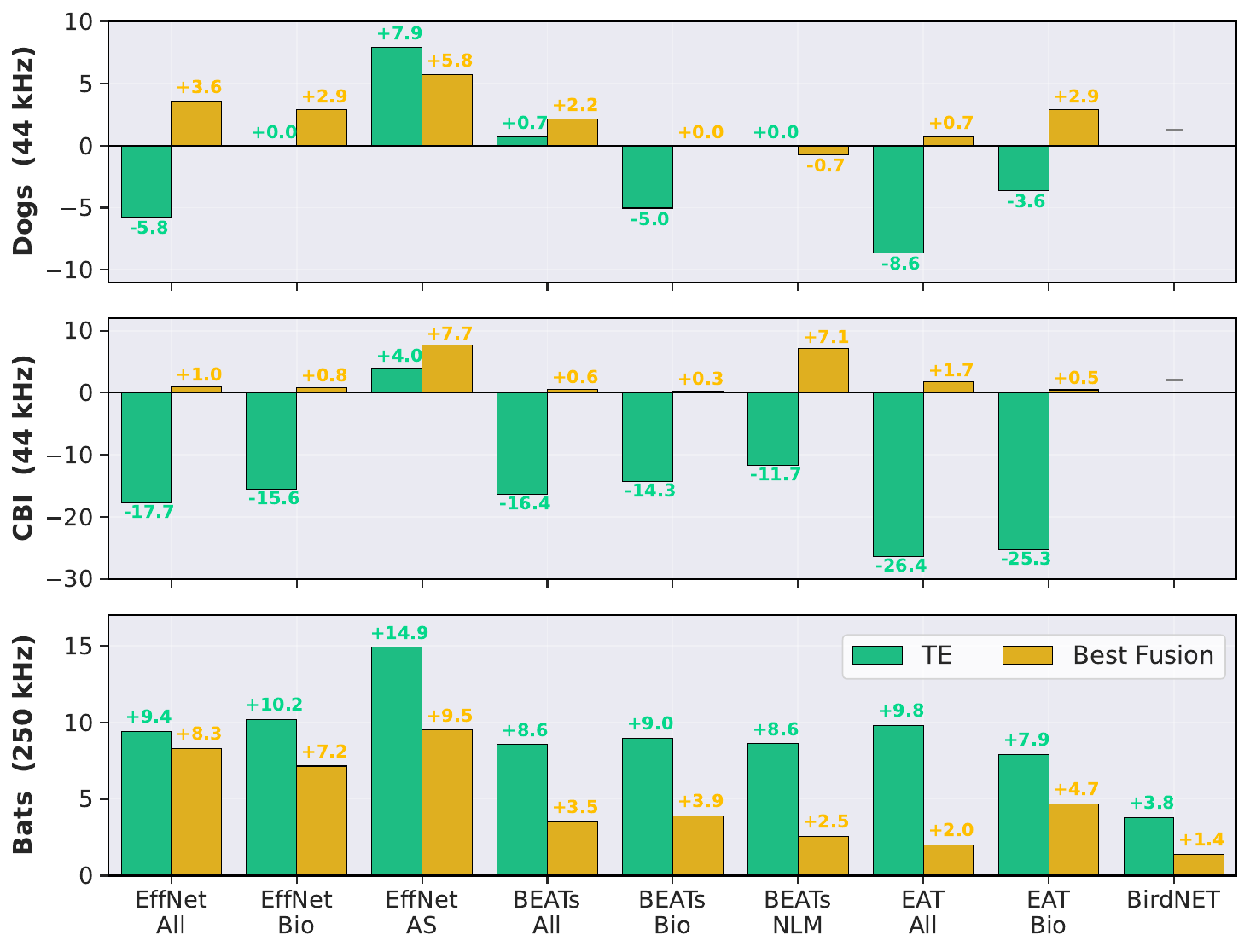}
  \vspace*{-0.6cm}
  \caption{\textit{Test} classification accuracy gain over baseband [\%].}
  \label{fig:bb_te_bars}
\end{figure}

\Cref{fig:bb_te_bars} visualizes the results as a bar plot, showing the gain of TE and the best fusion (per model) over the BB. We can observe that the fusion methods improve on the BB for almost all models across all datasets. Conversely, in the case of Dogs and CBI, we can clearly see that TE performs considerably worse than the BB across most models. Both TE and fusion provide significant gain for Bats over the BB, demonstrating that for species vocalizing at high frequencies, the additional spectral content beyond the baseband is highly informative, and leveraging it through multi-band encoding can clearly benefit bioacoustics tasks. In this case, TE proves to be even more robust, consistently outperforming fusion across all models.

\begin{table*}[ht]
\caption{Test accuracy [\%] by model across methods. \textbf{Bold} and \underline{underline} denote best and second-best performances, respectively.}
\centering
\label{tab:results_full}
\resizebox{\textwidth}{!}{%
\begin{tabular}{llccccccccc}
\toprule
Dataset & $\mathcal{F}$ & EffNet-All & EffNet-Bio & EffNet-AS & BEATs-All & BEATs-Bio & BEATs-NLM & EAT-All & EAT-Bio & BirdNET \\
& (SR)  & (16 kHZ) & (16 kHZ) & (16 kHZ) & (16 kHZ) & (16 kHZ) & (16 kHZ) & (16 kHZ) & (16 kHZ) & (48 kHZ) \\
\midrule
\multirow{7}{*}{\textbf{Dogs}} & BB & 92.81 & 89.21 & 77.70 & 87.05 & \textbf{92.81} & \textbf{87.05} & \underline{76.26} & 86.33 & 89.21 \\
 & TE & 87.05 & 89.21 & \textbf{85.61} & 87.77 & 87.77 & \textbf{87.05} & 67.63 & 82.73 & -- \\
 & MP  & 89.21 & \textbf{92.09} & 72.66 & \underline{88.49} & 91.37 & 81.29 & 52.52 & \textbf{89.21} & -- \\
 & GP  & \textbf{96.40} & 87.05 & 77.70 & 84.89 & 90.65 & \underline{86.33} & 69.06 & \underline{88.49} & -- \\
 & MoE & 92.81 & 89.21 & \underline{83.45} & 87.77 & \textbf{92.81} & 84.17 & 75.54 & 86.33 & -- \\
 & HYB & 89.93 & \underline{91.37} & 71.94 & 87.77 & \textbf{92.81} & 76.26 & 56.83 & \textbf{89.21} & -- \\
 & SA  & \underline{93.53} & 87.77 & 65.47 & \textbf{89.21} & \underline{92.09} & 80.58 & \textbf{76.98} & \textbf{89.21} & -- \\
\midrule
\multirow{7}{*}{\textbf{CBI}} & BB & 73.81 & 75.77 & 21.82 & 78.98 & 80.94 & \underline{56.55} & 67.65 & \underline{78.34} & 69.25\\
 & TE  & 56.13 & 60.19 & 25.80 & 62.62 & 66.66 & 44.83 & 41.27 & 53.01 & -- \\
 & MP  & 65.22 & 67.57 & 19.17 & 72.71 & 75.08 & 48.65 & 58.31 & 71.49 & -- \\
 & GP  & \underline{74.45} & \underline{75.99} & \underline{26.13} & 77.93 & 79.97 & 56.10 & 67.43 & 78.18 & -- \\
 & MoE & \textbf{74.78} & \textbf{76.60} & \textbf{29.50} & \underline{79.48} & \underline{80.99} & \underline{56.55} & \underline{68.29} & \textbf{78.81} & -- \\
 & HYB & 64.97 & 67.46 & 17.51 & 72.54 & 75.08 & 49.34 & 58.18 & 71.46 & -- \\
 & SA  & 71.41 & 69.75 & 21.71 & \textbf{79.53} & \textbf{81.24} & \textbf{63.67} & \textbf{69.39} & 77.76 & -- \\
\midrule
\multirow{7}{*}{\textbf{Bats}} & BB & 60.85 & 60.75 & 51.30 & 65.15 & 64.40 & 65.45 & 50.75 & 62.85 & 70.20\\
 & TE & \textbf{70.25} & \textbf{70.95} & \textbf{66.20} & \textbf{73.70} & \textbf{73.35} & \textbf{74.05} & \textbf{60.55} & \textbf{70.75} & \textbf{74.00} \\
 & MP & 52.45 & 50.45 & 45.30 & 50.50 & 49.10 & 45.55 & 29.30 & 47.20 & 62.50 \\
 & GP & 60.35 & 60.35 & 51.65 & 61.45 & 60.90 & 57.55 & 38.60 & 60.15 & 68.30 \\
 & MoE & \underline{69.15} & \underline{67.90} & \underline{60.80} & \underline{68.65} & \underline{68.30} & \underline{68.00} & \underline{52.75} & 66.30 & 71.20 \\
 & HYB & 52.10 & 49.85 & 44.20 & 48.80 & 49.85 & 45.85 & 31.10 & 46.05 & 63.45 \\
 & SA & 63.55 & 62.50 & 53.85 & 67.80 & 67.20 & 62.55 & 47.65 & \underline{67.55} & \underline{71.60} \\
\bottomrule
\end{tabular}}
\end{table*}

\Cref{tab:results_full} presents the full results across all baselines and fusion methods. Among fusion strategies, MoE outperforms the other methods in the majority of cases, especially dominating for CBI and Bats. This indicates that classifying each band embedding independently and then fusing the logits is more effective than combining them before classification, as in GP. SA also performs well, particularly for BEATs and EATs on CBI.

Note that since BirdNET operates at 48 kHz, Dogs and CBI calls fall entirely within its native BW and are only evaluated on the BB. Since bat calls extend beyond 24 kHz, we evaluate them with TE and fusion. For Dogs and CBI, we can observe that most of the 16 kHz models combined with fusion strategies are highly competitive and often outperform BirdNET's native 24 kHz BB, highlighting the effectiveness of our multi-band approach. Furthermore, on Bats, BirdNET achieves the highest score across all fusion methods, outperforming every corresponding 16 kHz model, and demonstrating the adaptability of this multi-band framework to other models with higher SR. 

Finally, EffNet-All and Bio yield comparable and notably higher fusion scores than EAT-All for all datasets, proving that the decorrelated band features (\cref{fig:bandwise_bats}) can provide complementary information that can be fused into salient representations.

\section{Conclusion}
\label{sec:conclusion}
This paper addressed a core limitation in computational bioacoustics: the loss of spectral information above the pre-training bandwidth of speech and audio models, typically fixed to 8 kHz. We investigated an adaptive multi-band fusion encoding framework that leverages the full spectrum of bioacoustic vocalizations, compared to baseband and time-expansion baselines. Similarity analyses showed that the framework can produce decorrelated band embeddings for some models and more class-discriminative fused representations on certain datasets. Using a linear classifier on these representations yielded improvements over the conventional baseband baseline across nearly all fusion strategies and encoder models, and substantially outperformed time-expansion on two datasets. The multi-band method also matched, and in some cases exceeded, the baseband performance of a model pre-trained at a higher bandwidth. Applying the framework to this higher-bandwidth model further produced representations that consistently outperformed those from lower-bandwidth models across all fusion strategies.

Overall, these results highlight the usefulness of this simple approach to overcome a key issue in processing and encoding animal recordings. Future work could explore more sophisticated multi-band techniques, such as overlapping or variable-width bands, to further improve robustness of bioacoustic representations for decoding non-human animal vocalizations.

\clearpage
\printbibliography
\clearpage

\end{document}